%% file: ms.tex
\documentclass[10pt,twocolumn,letterpaper]{article}
\usepackage{cvpr}

\usepackage[T1]{fontenc}
\usepackage[utf8]{inputenc}
\usepackage[english]{babel}
\usepackage{newtxtext}
\usepackage[vvarbb]{newtxmath}
\usepackage{courier}
\usepackage{textcomp}
\usepackage{microtype}
\usepackage{hyphenat}
\usepackage[all]{nowidow}
\usepackage{enumitem}

\usepackage{amsmath}
\usepackage{amssymb}
\usepackage{amsfonts}
\usepackage{bm}
\usepackage{xfrac}
\usepackage{siunitx}

\usepackage{booktabs}
\usepackage{tabulary}

\usepackage{epsfig}
\usepackage{graphicx}
\usepackage[labelsep=period,labelfont=bf,font={small}]{caption}
\usepackage{subcaption}
\usepackage{dblfloatfix} %
\usepackage{pifont}
\usepackage{lipsum}
\usepackage[export]{adjustbox}
\usepackage{float}

\usepackage{algorithm}
\usepackage{algpseudocode}
\usepackage{algorithmicx}

\usepackage{xcolor}
\usepackage{soul}

\usepackage[pagebackref=true,breaklinks=true,letterpaper=true,colorlinks,bookmarks=false]{hyperref}
\usepackage{cleveref}

\graphicspath{{./images/}}

\newcommand{\comment}[1]{\ignorespaces}

\newcommand{\overbar}[1]{\mkern 1.5mu\overline{\mkern-1.5mu#1\mkern-1.5mu}\mkern 1.5mu}

\def\*#1{\bm{#1}}  %
\DeclareMathAlphabet\mathcalbf{OMS}{cmsy}{b}{n}
\newcommand{\btheta}{\bm{\theta}}

\DeclareMathOperator*{\argmin}{arg\,min}

\AtBeginDocument{
 \abovedisplayskip=10pt plus 3pt minus 6pt
 \abovedisplayshortskip=0pt plus 3pt
 \belowdisplayskip=10pt plus 3pt minus 6pt
 \belowdisplayshortskip=7pt plus 3pt minus 4pt
}

\definecolor{highlightcolor}{RGB}{255,237,191}
\definecolor{textboxcolor}{RGB}{155,155,155}
\definecolor{paperboxcolor}{RGB}{204,148,73}
\definecolor{highlightred}{RGB}{224,103,103}
\definecolor{greencustom}{rgb}{0.0, 0.5, 0.0}
\definecolor{lightgray}{rgb}{0.83, 0.83, 0.83}

\newcommand{\ra}[1]{\renewcommand{\arraystretch}{#1}}

\newcommand{\ms}{\si{\meter\per\second}}

\newcommand{\kgmm}{\si{\kilo\gram\per\square\meter}}

\cvprfinalcopy

\title{Go with the Flow: Perception-refined Physics Simulation}

\author{Tom F.~H. Runia \hspace{1.4em} Kirill Gavrilyuk \hspace{1.4em} Cees G.~M. Snoek \hspace{1.4em} Arnold W.~M. Smeulders \\
    QUVA Deep Vision Lab, University of Amsterdam\\
    {\tt\small \{runia,kgavrilyuk,cgmsnoek,a.w.m.smeulders\}@uva.nl}
}

\hypersetup{pdfcreator={Tom F.H. Runia, Kirill Gavrilyuk, Cees G.M. Snoek and Arnold W.M. Smeulders}}
\hypersetup{pdfproducer={Tom F.H. Runia}}
\hypersetup{pdftitle={Go with the Flow: Perception-refined Physics Simulation}}
\hypersetup{pdfsubject={Computer Vision}}

\begin{document}

\maketitle
\thispagestyle{empty}

\begin{abstract}
  \vspace{-3mm}
  For many of the physical phenomena around us, we have developed sophisticated models explaining their behavior. Nevertheless, inferring specifics from visual observations is challenging due to the high number of causally underlying physical parameters -- including material properties and external forces. This paper addresses the problem of inferring such latent physical properties from observations. Our solution is an iterative refinement procedure with simulation at its core. The algorithm gradually updates the physical model parameters by running a simulation of the observed phenomenon and comparing the current simulation to a real-world observation. The physical similarity is computed using an embedding function that maps physically similar examples to nearby points. As a tangible example, we concentrate on flags curling in the wind -- a seemingly simple phenomenon but physically highly involved. Based on its underlying physical model and visual manifestation, we propose an instantiation of the embedding function. For this mapping, modeled as a deep network, we introduce a spectral decomposition layer that decomposes a video volume into its temporal spectral power and corresponding frequencies. In experiments, we demonstrate our method's ability to recover intrinsic and extrinsic physical parameters from both simulated and real-world video.
\end{abstract}

\input{01_introduction}

\input{02_related_work}
\input{03a_method_main}

\input{03b_method_similarity}

\input{03c_method_optimization}

\input{04a_flags_physics}

\input{04b_flags_network}

\input{04c_flags_datasets}

\input{05_experiments}

\input{06_conclusion}

{\small
  \bibliographystyle{ieee_fullname}
  \bibliography{ms}
}

\end{document}

%% file: 01_introduction.tex
\vspace{-5mm}

\section{Introduction}
\label{sec:introduction}

There is substantial evidence \cite{hegarty2004mechanical,craik1967nature} that humans run mental models to predict physical phenomena. We predict the trajectory of objects in mid-air, anticipate the pouring of milk in a glass and estimate the velocity at which an object slides down a ramp. In analogy, simulation models usually optimize their parameters by performing trial runs and selecting the best. In this paper, we aim to optimize physical simulations by aligning them with visual observations.

\begin{figure}
	\centering
    \includegraphics[width=\columnwidth]{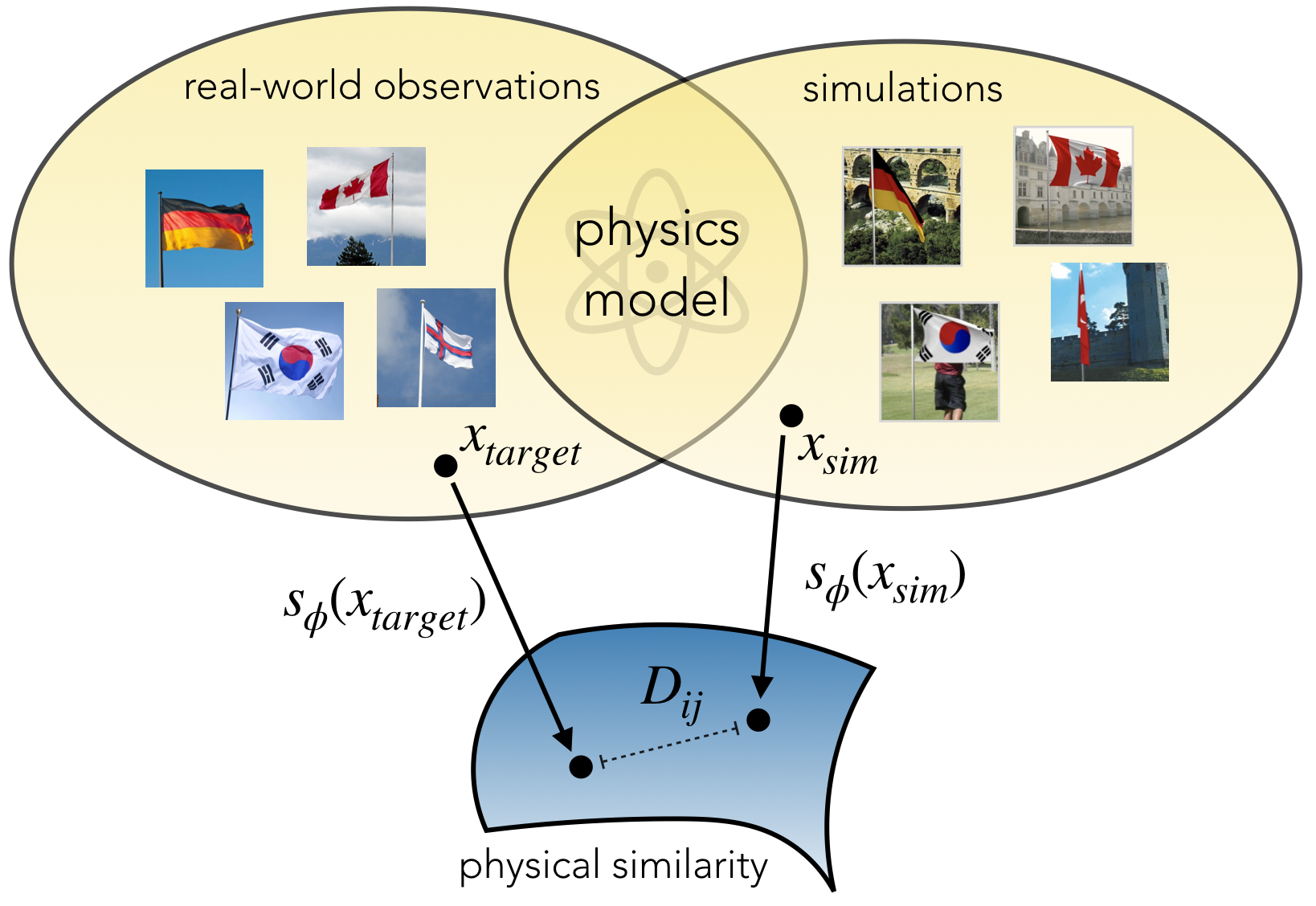}
    \vspace{-5mm}
    \caption{We propose intrinsic and extrinsic physical parameter refinement based on real-world perception. To guide the optimization, we determine physical similarity between simulation and real-world using an embedding function that is modeled as a deep network. \label{fig:intro-figure}}
    \vspace{-3mm}
\end{figure}

The performance of a model is determined by its parameters. The best set of parameters may be defined by (1) a human observer selecting the run that is the most realistic, or (2) the model parameters may be measured and inserted into the model \cite{wang2011data}. We propose a method for optimizing simulation parameters without the reliance on either of these two existing solutions. Specifically, we introduce an iterative refinement procedure to build physically realistic models by aligning them with real-world observations.

The task is challenging, as physical models tend to have high numbers of unknown parameters and bear intricate coupling of intrinsic and external forces. A key component is how to compare the simulated data with real-world data. One solution would be to align both visual appearances directly. However, we maintain that when an object's appearance is formed by external physical forces, physical realism is a better principle to align the simulations with real-world data.

At the core of our method is a physics simulation engine with unknown parameters $\btheta$ to be optimized. The outcome of a simulation (\eg 3D meshes, points clouds, flow vectors) is converted to the image space using a render engine. We then compare the simulated visual data with a real-world observation of the particular phenomenon. Accordingly, we propose to learn a \emph{physical similarity} metric from simulated examples, each with known physical parameters. In the learned embedding space, points with similar physical parameters will wind up close while dissimilar example pairs will be further away (\Cref{fig:intro-figure}). Guided by the physical similarity, the simulation's parameters are refined in each step. As a result, we obtain a complete computational solution for refining a physical model.

\begin{figure}[t]
    \centering
    \includegraphics[width=\columnwidth,trim={0 20cm 16.5cm 0},clip]{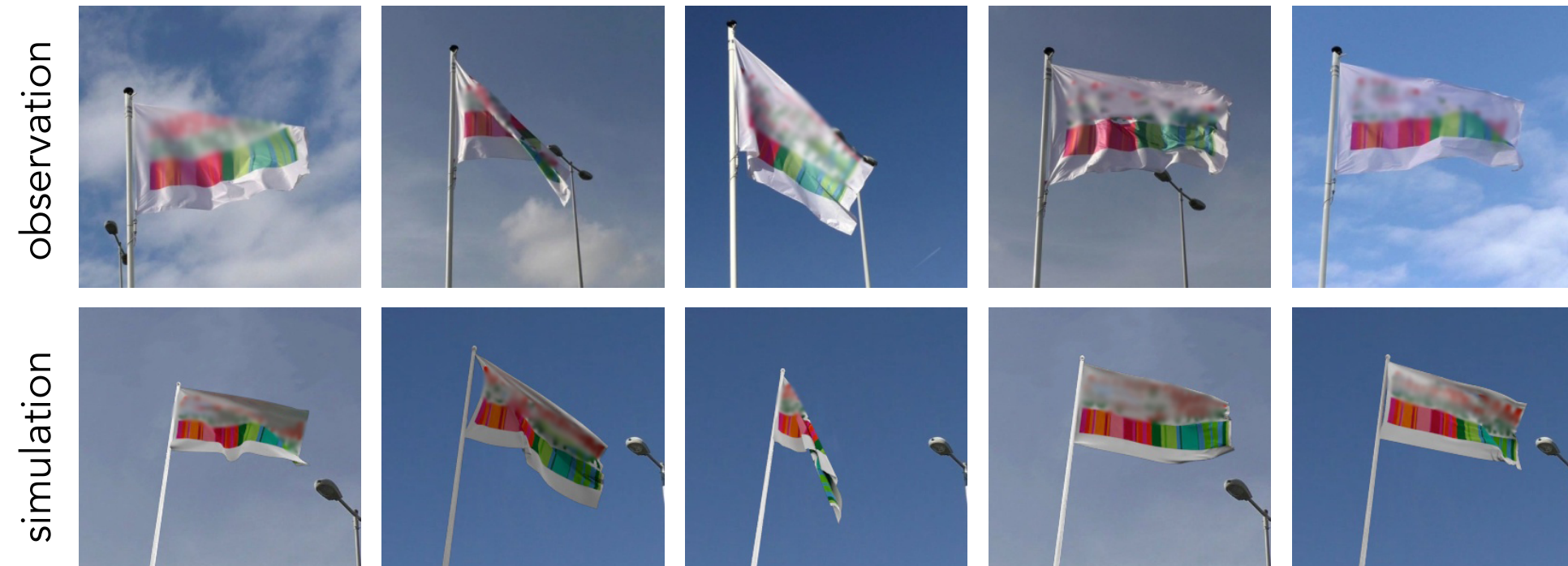}
    \vspace{-5mm}
    \caption{\emph{Top row:} random still images from our video recordings of real flags. \emph{Bottom row:} random still images produced by our flag simulation pipeline (\Cref{fig:method-overview}). The simulated flags' dynamics closely correspond to real-world flag dynamics as our method estimates intrinsic and extrinsic physical model parameters. For review purposes we have obfuscated text on the flag.}\label{fig:flagsim-comparison-flagreal}
    \vspace{-5mm}
\end{figure}

We have selected a flag as it curls in the wind's airflow as our leading example (\Cref{fig:flagsim-comparison-flagreal}). Its physical dynamics are determined by the cloth's intrinsic material properties as well as the external wind force. Untangling the dynamics of flags is challenging due to the involved nature of the air-cloth interaction. Specifically, a flag exerts inertial and elastic forces on the surrounding air, while the air acts on the fabric through pressure and viscosity \cite{huang2010three}. As we seek to infer both the cloth's intrinsic material properties and the external wind force, our physical model couples a non-linear cloth model \cite{wang2011data} with external wind force \cite{wejchert1991animation}. While the flapping motion is most familiar for its self-sustained oscillation in flags, similar models emerge in paper production, energy harvesting and the undulatory propulsion of fish \cite{shelley2011flapping,huang2010three}. Other physical systems we could consider include fire, smoke, foam, fluid dynamics and mechanical problems.

This paper makes the following contributions: (1) We propose the perception-based refinement of both intrinsic and extrinsic physical model parameters. This is achieved through a simulation engine that implements the physical model. (2) To guide the parameter optimization in each step, we quantify the correspondence between simulation and real-world using a learned embedding function that maps physically similar examples to nearby points. (3) To demonstrate our method's physical parameter refinement, we consider the case of a flag in the wind. (4) In this light, we introduce a specification of the embedding function. It is modeled as a deep network and adopts our new \emph{spectral decomposition layer} that extracts temporal spectral information while maintaining spatial structure. To evaluate our method, we collect real-world video of flags, together with the ground-truth wind speed gauged using an anemometer.

\begin{figure*}[t]
    \centering
    \includegraphics[width=\textwidth,trim={0 18.5cm 7.7cm 0},clip]{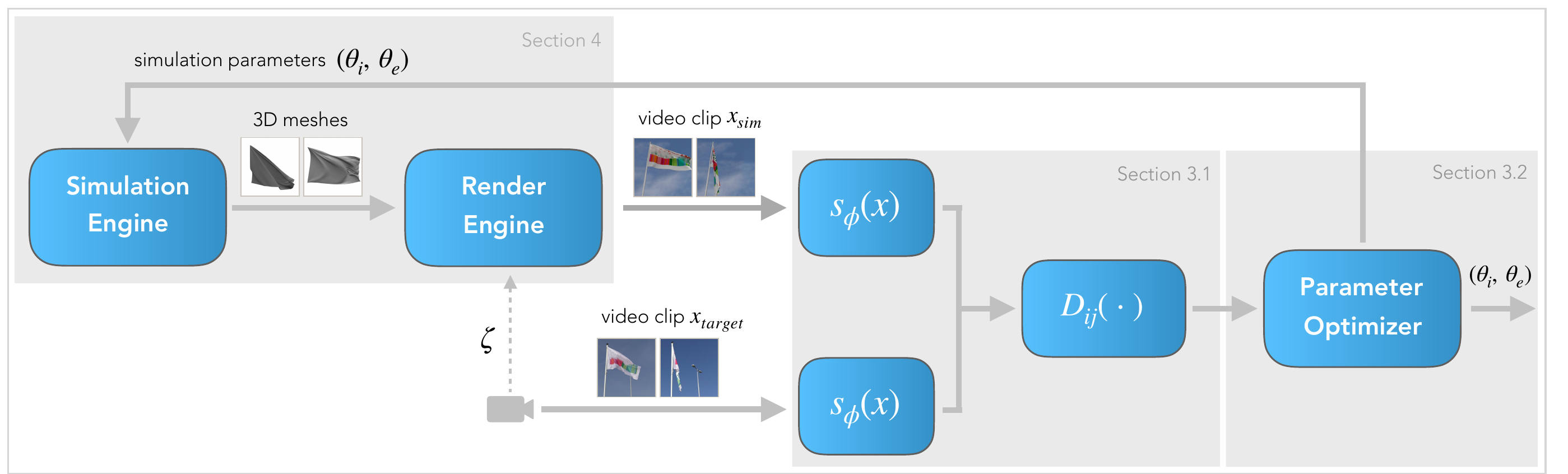}
    \vspace{-6mm}
    \caption{We propose perception-based refinement of physics simulations. Given an observation of a real-world physical phenomenon, here represented as video clip $\*x_{\text{target}}$, our algorithm fits the causally underlying parameters responsible for the phenomenon's manifestation. Central is a simulation engine implementing the physical model, parametrized by intrinsic material properties $\btheta_i$ and the characterization of external forces $\btheta_e$. A render engine, with render parameters $\*\zeta$, maps the simulator's output to the image space producing video clip $\*x_{\text{sim}}$. Using an embedding function $s_{\phi}(\*x)$ both real and simulated examples are mapped to a manifold on which physically similar examples are assigned to nearby points. To quantify correspondence between both clips we evaluate a distance metric $D_{ij}(\cdot,\cdot)$ in the embedding space. The distance score is fed into an optimization module that adjusts the physical parameters $\btheta$ towards the real-world observation.} \label{fig:method-overview}
    \vspace{-2mm}
\end{figure*}

%% file: 02_related_work.tex
\section{Related Work}
\label{sec:related-work}

 Previous work has recovered physical properties by perceiving real-world objects or phenomena -- including material properties \cite{davis2015visual}, cloth stiffness and bending parameters \cite{bouman2013estimating,yang2017learning}, mechanical features \cite{wu2015galileo,mottaghi2016newtonian,mottaghi2016happens,li2016fall}, fluid characteristics \cite{wu2016physics,spencer2004water,sakaino2008fluid} and surface properties \cite{meka2018lime}. The primary focus of the existing literature has been on estimating intrinsic material properties from visual input. In this work, we consider the more complex scenario of jointly estimating intrinsic material properties and extrinsic forces from a video.

In computer vision, the visual appearance of structures under the influence of wind has been studied before, including the oscillation of tree branches \cite{xue2018seeing,sun2003video}, water surfaces \cite{spencer2004water}, and hanging cloth \cite{bouman2013estimating,yang2017learning}. However, until recently the visual appearance of flags was not studied in computer vision. Recently, independently from our own work, Cardona \etal \cite{cardona2019seeing} proposed the use of convolutional neural networks to infer the wind speed from short videos. While the curling of a flag may be perceived as simple at first, its motion is highly complex. Fortunately, its dynamics are an important and well-studied topic in the field of fluid-body interactions \cite{shelley2011flapping,taneda1968waving,tian2013role}. In computer graphics, cloth simulation has been studied by numerous works \cite{provot1995deformation,wang2011data,narain2012adaptive,bridson2005simulation}. Based on both physics and graphics literature, we settle on Wang's \etal \cite{wang2011data} cloth model efficiently implemented in ArcSim \cite{narain2012adaptive}.

The visual appearance of flags is an instance of the more general topic of cloth appearance, which is widely studied in computer vision. Most successful approaches characterize cloth by its wrinkles, folds and silhouette \cite{bhat2003estimating,haddon1998shading,white2007capturing,yang2018physics}. The existing literature that recovers cloth material properties from video \cite{bhat2003estimating,bouman2013estimating,yang2017learning} is closest to our study on flags. Notably, Bouman \etal \cite{bouman2013estimating} use complex steerable pyramids to describe hanging cloth in a video, while Yang \etal \cite{yang2017learning} adopt a learning-based approach combining a convolutional network and recurrent network. We propose to characterize flags by the spatial distribution of temporal frequencies over the flag's surface. This is inspired by recent work in video recognition that extracts the spatial distribution of temporal frequencies \cite{lindeberg2018dense,runia2018repetition}. Unlike previous work, we adopt the Discrete Fourier Transform for its use inside a deep network layer. We compare our cloth frequency-based representations to Yang \etal \cite{yang2017learning} on the cloth dataset of Bouman \etal \cite{bouman2013estimating}.  

There have been a few works that connect the physical parameter recovery to computer simulations. Our approach shares similarity to the Monte Carlo-based parameter optimization of \cite{wu2015galileo} and the particle swarm refinement of clothing parameters from static images \cite{yang2018physics}. In particular, the work of \cite{yang2017learning} resembles ours as they infer garment properties from static images for the purpose of virtual clothing try-on. However, our work is different in an important aspect: while their work focuses on estimating intrinsic cloth properties in static equilibrium, we estimate both intrinsic and extrinsic physical parameters from a video.

%% file: 03a_method_main.tex
\section{Method}
\label{sec:method}

We consider the scenario in which we make an observation of some phenomena with a physical model explaining its manifestation available to us. Based on the perception of reality, our goal is to fit the $D_p$ unknown continuous parameters of the physical model $\btheta \in \mathbb{R}^{D_p}$, consisting of intrinsic parameters $\btheta_i$ and extrinsic parameters $\btheta_e$, through iterative refinement of a computer simulation that implements the physical phenomena at hand. In particular, we consider observations in the form of short video clips $\*x_{\text{target}} \in \mathbb{R}^{C\times N_t \times H \times W}$, with $C$ denoting the number of image channels and $N_t$ the number of $H \times W$ frames. In each iteration, a simulator runs with current model parameters $\btheta$ to produce some intermediate representation (\eg 3D meshes, point clouds or flow vectors), succeeded by a render engine with parameters $\*\zeta$ that yields a simulated video clip $\*x_{\text{sim}} \in \mathbb{R}^{C\times N_t \times H \times W}$. Our insight is that real-world observation and simulation can be compared in some embedding space using pairwise distance:
\begin{align}
    D_{i,j} = D\left(s_{\phi}(\*x_i), s_{\phi}(\*x_j)\right) : \mathbb{R}^{D_e}\times \mathbb{R}^{D_e} \rightarrow \mathbb{R}
    \label{eq:distance-function}
\end{align}
where $s_{\phi}\*(\*x) : \mathbb{R}^{C\times N_t \times H \times W} 
\rightarrow \mathbb{R}^{D_e}$ an embedding function parametrized by $\phi$ that maps the data manifold $\mathbb{R}^{C\times N_t \times H \times W}$ to some embedding manifold $\mathbb{R}^{D_e}$ on which physically similar examples should lie close. In each iteration, based on distance \eqref{eq:distance-function} between real and simulated instance, the physical model parameters are updated to maximize their correspondence. This procedure ends whenever the physical model parameters have been fitted accurately within some predefined tolerance, or when the evaluation budget is finished. The output is the optimal set of parameters $\btheta^*$ and corresponding simulation $\*x_{\text{sim}}^*$ of the real-world phenomenon. An overview of the proposed method is presented in \mbox{\Cref{fig:method-overview}}.

%% file: 03b_method_similarity.tex
\subsection{Physical Similarity}
\label{subsec:similarity-function}

For the parameter refinement to be successful, it is crucial to quantify the similarity between simulation $\*x_{\text{sim}}$ and real-world observation $\*x_{\text{target}}$. The similarity function must reflect correspondence in physical dynamics between the two instances. The prerequisite is that the physical model must describe the phenomenon's behavior at the scale that coincides with the observational scale. For example, the quantum mechanical understanding of a pendulum will be less meaningful than its formulation in classical mechanics when capturing its appearance using a regular video camera. 

Given the physical model and its implementation as a simulation engine, we generate a collection of simulations with its parameters randomly sampled from some predefined search space. For each of these simulated representations of the physical phenomenon, we use a 3D render engine to generate multiple video clips $\*x_{\text{sim},}^i$ with different render parameters $\*\zeta^i$. As a result, we obtain a dataset with multiple renders for each simulation instance. Given this dataset we propose the following training strategy to learn an embedding function. 

We employ a \emph{contrastive loss} \cite{hadsell2006dimensionality} and consider positive example pairs to be rendered video clips originating from the same simulation (\ie sharing physical parameters) while negative example pairs belong to different simulations. Both video clips of an example pair are mapped to the embedding space through $s_{\phi}(\*x)$ in Siamese fashion \cite{bromley1994signature}. In the embedding space, the physical similarity will be evaluated using the squared Euclidean distance: $D_{i,j} = D\left(s_{\phi}(\*x_i), s_{\phi}(\*x_j)\right) = \Vert s_{\phi}(\*x_i) - s_{\phi}(\*x_j) \Vert_2^2$. After training on the simulated dataset using the described sampling strategy, the learned embedding function and distance function can be used for quantifying correspondence between simulations and real-world observations.

%% file: 03c_method_optimization.tex
\subsection{Simulation Parameter Optimization}
\label{subsec:parameter-optimization}

We observe that updating the physical model parameters in our algorithm (\Cref{fig:method-overview}) resembles the problem of hyperparameter optimization \cite{snoek2012practical,bergstra2012random}. In light of this correspondence, our collection of model parameters is analogous to the hyperparameters involved by training deep neural networks (\eg learning rate, weight decay, dropout). Formally, we seek to find the global optimum of physical parameters: 
\begin{align}
    \btheta^* = \argmin_{\btheta} \, D\left(s_{\phi}(\*x_\text{target}), s_{\phi}(\*x_\text{sim}(\btheta))\right),
    \label{eq:hyperparam-optimum}
\end{align}
where the target example is fixed and the simulated example depends on the current set of physical parameters $\btheta$. Adjusting the parameters $\btheta$ at each iteration is challenging as we can make no parametric assumptions on \eqref{eq:hyperparam-optimum} as function of $\btheta$, we have no access to the gradient and its evaluation is costly as it involves running a potentially expensive physics simulation. Our goal is, therefore, to estimate the global minimum with as few evaluations as possible. Considering this, we adopt Bayesian optimization \cite{snoek2012practical} for updating parameters $\btheta$. Its philosophy is to leverage all available information from previous observations of \eqref{eq:hyperparam-optimum} and not only use local gradient information. We treat the optimization as-is and use a modified implementation of Spearmint \cite{snoek2012practical} with the Mat\'ern52 kernel and improved initialization of the acquisition function \cite{oh2018bock}. Note that the embedding function $s_{\phi}(\*x)$ is fixed throughout this optimization.

%% file: 04a_flags_physics.tex
\begin{figure}
	\centering
    \includegraphics[width=\columnwidth,trim={0 16.5cm 20.5cm 0},clip]{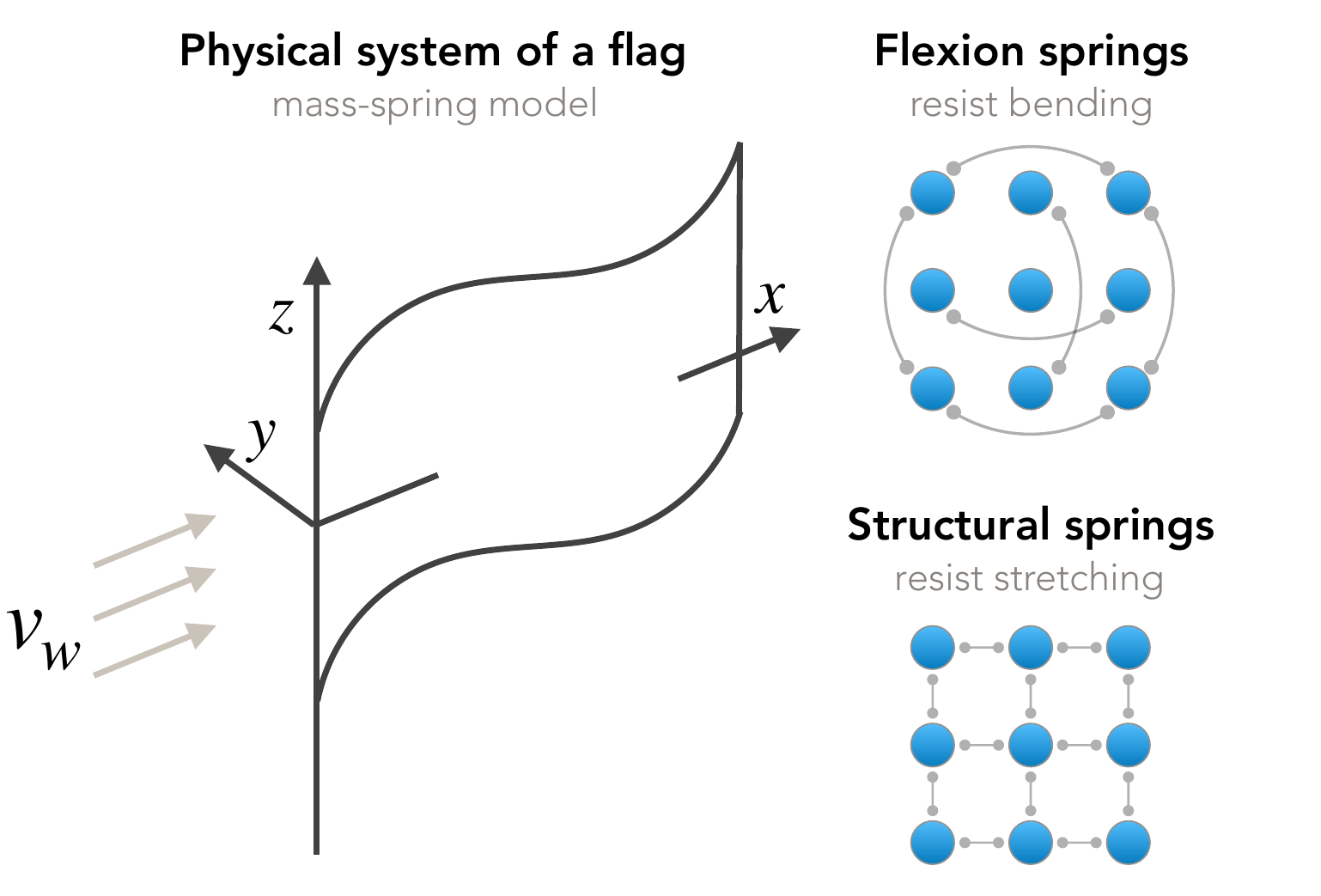}
    \vspace{-7mm}
    \caption{\emph{Left:} the flag's fabric is treated as a mass-spring model in which a dense grid of point masses is inter-connected with multiple springs. \emph{Right:} most significantly, the bending and stretching springs determine the materials behavior. Flexion springs act over shared edges whereas structural springs connect to direct neighbors. \label{fig:flag-illustration}}
    \vspace{-5mm}
\end{figure}

\section{Physics, Simulation and Appearance of Flags}
\label{sec:physical-model}

Up until now, we have discussed the proposed method in its most general terms. Here, we will transition to a specific instantiation of a physical phenomenon. Specifically, we will consider the challenging case of \emph{a flag curling in the wind}. By doing so, we confine the parameters $\btheta$ and embedding function $s_{\phi}(\*x)$. However, we emphasize that the algorithm generalizes to any physical phenomena given that the underlying physical model is known and a perceptual distance function can be defined.

\subsection{Physical Model}
\label{subsec:physical-model-of-flags}

The physical understanding of cloth and its interaction with external forces has been assimilated by the computer graphics community. Most successful methods treat cloth as a mass-spring model (\Cref{fig:flag-illustration}): a dense grid of point masses organized in a planar structure, inter-connected with different types of springs which properties determine the fabric's behavior \cite{baraff1998large,provot1995deformation,wang2011data,baraff2003untangling,narain2012adaptive}. We adopt Wang's \etal \cite{wang2011data} non-linear and anisotropic mass-spring model for cloth. This model uses a piecewise linear bending and stretching models. The stretching model is a generalization of Hooke's law for continuous media \cite{slaughter2012linearized}. In the context of flags, the stretching properties are of minimal relevance as flags are typically made of strong weather resistant materials such as polyester and nylon. Consequently, we emphasize on the cloth's bending model \cite{wang2011data} and external forces \cite{wejchert1991animation}.

\vspace{2mm}

\noindent \textbf{Bending Model ($\btheta_i$).} The bending model is based on the linear bending force equation first proposed in \cite{bridson2005simulation}. The model formulates the elastic bending force $\*{F}_e$ over triangular meshes sharing an edge (\Cref{fig:flag-illustration}). For two triangles separated by the dihedral angle $\varphi$, the bending force reads:
\begin{align}
    \*{F}_e = k_e \sin(\varphi / 2)(N_1 + N_2)^{-1} \vert \*{E} \vert \*{u},
    \label{eq:bending-equation}
\end{align}
where $k_e$ is the material dependent bending stiffness, $N_1, N_2$ are the weighted surface normals of the two triangles, $\*{E}$ represents the edge vector and $\*{u}$ is the bending mode (see Figure~1 in \cite{bridson2005simulation}). The bending stiffness $k_e$ is non-linearly related to the dihedral angle $\varphi$. This is realized by treating $k^e$ as piecewise linear function of the reparametrization $\alpha = \sin(\varphi/2)(N_1+N_2)^{-1}$. After this reparametrization, for a certain fabric, the parameter space is sampled for $N_b$ angles yielding a total of $3N_b$ parameters across the three directions. Wang \etal \cite{wang2011data} empirically found that $5$ measurements are sufficient for most fabrics, producing $15$ bending parameters. 

\vspace{2mm}

\noindent \textbf{External Forces ($\btheta_e$).} For the dynamics of a waving flag, we consider two external forces acting upon its planar surface. First, the Earth's gravitational acceleration (\mbox{$\*F_g = m\*a_g$}) naturally pushes down the fabric. The total mass is defined by the flag's area weight $\rho_A$ multiplied by surface area. More interestingly, we consider a flag in a constant wind field. Again, modeling the flag as a grid of point masses, the drag force on each mass is stipulated by Stokes's equation \cite{batchelor1967introduction}: $\*F_d = 6 \pi R \eta \*v_w$, where $R$ the radius of a spherical particle, $\eta$ the air's dynamic viscosity and $\*v_w$ the wind velocity. By all means, this is a simplification of reality. Our model ignores terms associated with the Reynolds number (such as the flag's drag coefficient), which will also affect a real flag's dynamics. However, it appears that the model is accurate enough to cover the spectrum of flag dynamics. %

\begin{table}
    \centering
    \caption{The parameters $\btheta = (\btheta_i, \btheta_e)$ for the physical model of a flag curling in the wind. These parameters are refined during the optimization. The bending parameters $\overbar{k}_e$ correspond to the ``Camel Ponte Roma'' base material from \cite{wang2011data}. \label{tab:parameter-search-space} }
    \ra{1.1}
    \small
    \vspace{-2mm}
    \scalebox{0.96}[1.0]{
        \begin{tabular}{llcl}
            \toprule
            & Parameter & Params & Search space \\ 
            \midrule
            $\theta_i$ & Bending stiffness  & $15$ & $k_e \in [10^{-1}\overbar{k}_e, 10\overbar{k}_e]$ \\ 
            $\theta_i$  & Fabric area weight & $1$  & $\rho_A \in [0.10, 0.17]$ \kgmm{} \\ 
            $\theta_e$ & Wind velocity      & $1$  & $v_w \in [0,10]$ \ms{} \\ 
            \bottomrule
        \end{tabular}
    }
    \vspace{-3mm}
\end{table}

\subsection{Simulation Engine}
\label{subsec:simulation-engine}

We employ the ArcSim simulation engine \cite{narain2012adaptive} which implements the physical model described in \Cref{subsec:physical-model-of-flags}. On top of the physical model, the simulator incorporates anisotropic remeshing to improve detail in densely wrinkled regions while coarsening flat regions. As input, the simulator expects the cloth's initial mesh, its material properties and the configuration of external forces. At each time step, the engine solves the system for implicit time integration using a sparse Cholesky-based solver. This produces a sequence of 3D cloth meshes based on the physical properties of the scene. As our goal is to define a similarity function in image space between simulation and a real-world observation, we pass the sequence of meshes through a 3D render engine \cite{blender2018}. Given render parameters $\*\zeta$ comprising of camera position, scene geometry, lighting conditions and the flag's visual texture, the renderer produces a simulated video clip ($\*x_{\text{sim}}$) which we can compare directly to the real-world observation ($\*x_{\text{target}}$). We emphasize that our focus is neither on inferring render parameters $\*\zeta$ from the real-world observation nor on attaining visual realism for our renders.

\vspace{2mm}

\noindent \textbf{Parameter Search Space ($\btheta_i, \btheta_e$).} The ArcSim simulator \cite{narain2012adaptive} operates in metric units, enabling convenient comparison with real-world dynamics. As the base material for all experiments, we use ``Camel Ponte Roma'' from \cite{wang2011data}. Made of $60\%$ polyester and $40\%$ nylon, this material closely resembles the fabrics widely used for flags \cite{wang2011data}. The fabric's bending, stretching, and area weight was accurately measured in a mechanical setup by the authors. We adopt and fix their stretching parameters and use the bending stiffness and area weight as initialization for our flag material. Specifically, using their respective parameters we confine a search space that is used during our parameter refinement. We determine $\rho_A \sim \text{Uniform}(0.10, 0.17)$~\kgmm{} after consulting various flag materials at online retailers. And, we restrict the range of the bending stiffness coefficients by multiplying the base material's $\overbar{k}_e$ in \eqref{eq:bending-equation} by $10^{-1}$ and $10$ to obtain the most flexible and stiffest material respectively. As the bending coefficients have a complex effect on the cloth's appearance, we independently optimize the $15$ bending coefficients instead of only tuning the one-dimensional multiplier. The full parameter search space is listed in \Cref{tab:parameter-search-space}. %

%% file: 04b_flags_network.tex
\subsection{Spectral Decomposition Network}
\label{subsec:spectral-decomposition-network}

We seek a representation that can encode a flag's characteristics such as the high-frequent streamwise waves towards the trailing edge, the number of nodes in the flag's fabric, violent flapping at the trailing edge, rolling motion of the upper corner, sagging down of the flag and its silhouette \cite{shelley2011flapping,taneda1968waving,eloy2008aeroelastic}. This suggests, modeling the \emph{spatial distribution of temporal spectral power} over the flag's surface. Together with direction awareness, this effectively encodes horizontal traveling waves and rolling motion of the upper corner. 

\vspace{2mm}

\begin{algorithm}[b]
    \caption{Spectral Decomposition Layer}
    \label{alg:spectral-decomposition-layer}
    \begin{algorithmic}[1] %

        \State \textbf{Input.} Video tensor $\*x$ of shape $[N_b, C, N_t, H, W]$
        \State \textbf{Input.} Number of frequency peaks to select, $k$
        \State \textbf{Output.} Decomposition of shape $[N_b,2kC,H,W]$
        \vspace{3mm}

        \Procedure{SpectralDecompositionLayer}{$\*x$}
            \State Reshape $\*x$ to $[N_bCHW, N_t]$ to obtain batch of signals
            \State Apply a Hanning window to signals
            \State Compute the DFT of signals using \eqref{eq:DFT}
            \State Compute periodogram of signals $I(\omega)$
            \State Select top-$k$ peaks of $I(\omega)$ and corresponding $\omega$'s
            \State $P \leftarrow$ top-$k$ peaks of $I(\omega)$ reshaped to $[N_b,kC,H,W]$
            \State $\Omega \leftarrow$ corresponding $\omega$'s reshaped to $[N_b,kC,H,W]$
            \State \textbf{return} $P, \Omega$
        \EndProcedure
    \end{algorithmic}
\end{algorithm}

\begin{figure*}
    \centering
    \includegraphics[width=0.95\textwidth,trim={0 6cm 19.5cm 0},clip]{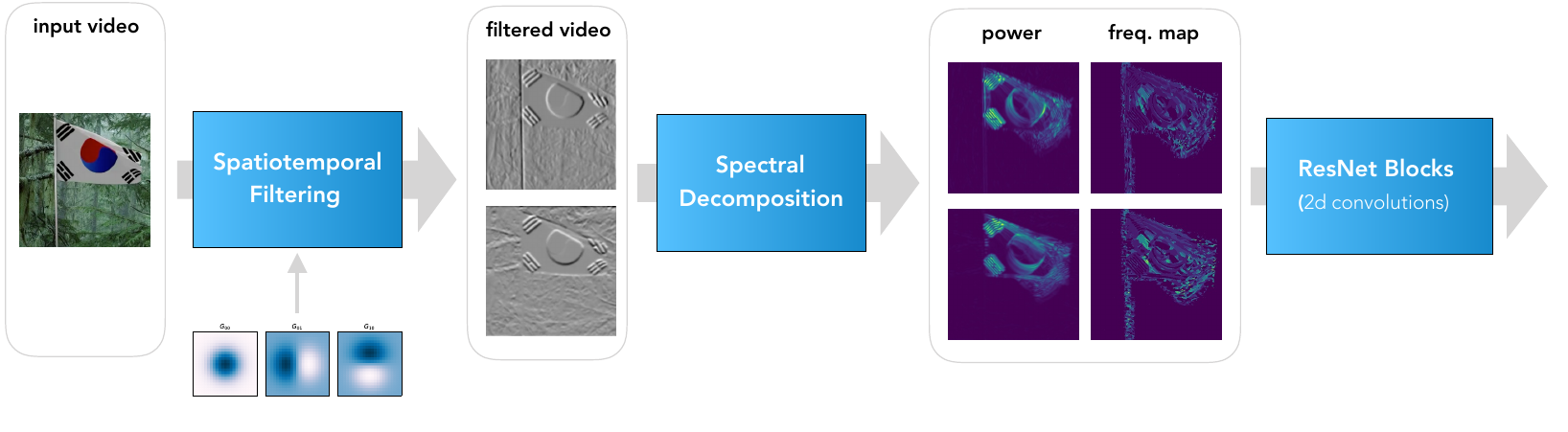}
    \vspace{-3mm}
    \caption{Overview of our network architecture for learning the physical correspondence between the simulation and real-world observation of dynamic flags. Given a 3D video volume as input, we first apply a $0^{\text{th}}$-order temporal Gaussian filter followed by two directional $1^{\text{st}}$-order Gaussian derivative filters and then spatially subsample both filtered video volumes by a factor two. The proposed spectral decomposition layer then applies the Fourier transform and selects the maximum power and corresponding frequencies densely for all spatial locations. This produces 2D multi-channel feature maps. Finally, a number of 2D convolution layers can process the feature maps to learn an embedding. 
    \label{fig:network-architecture}}
    \vspace{-4mm}
\end{figure*}

\noindent \textbf{Spectral Decomposition Layer.} We propose a spectral decomposition layer that distills temporal frequencies from a video. Specifically, we treat an input video volume as a collection of signals for each spatial position (\ie $H\times W$ signals) and map the signals into the frequency domain using the Discrete Fourier Transform (DFT) to estimate the videos' spatial distribution of temporal spectral power. The DFT maps a signal $f[n]$ for $n \in [0, N_t-1]$ into the frequency domain \cite{oppenheim1999discrete} as formalized by:
\begin{equation}
    F(j\omega) = \sum_{n=0}^{N_t-1} f[n] e^{-j \omega n T}.
    \label{eq:DFT}
\end{equation}
We proceed by mapping the DFT's complex output to a real-valued representation. The periodogram of a signal is a representation of its spectral power and is defined as $I({\omega}) = \frac{1}{N_t} \vert F(j\omega) \vert^2$ with $F(j \omega)$ as defined in \eqref{eq:DFT}. This provides the spectral power magnitude at each sampled frequency. To effectively reduce the dimensionality and emphasize on the videos' discriminative frequencies, we select the top-$k$ strongest frequencies and corresponding spectral power from the periodogram. Given a signal of arbitrary length, this produces $k$ pairs containing $I(\omega_{\max_i})$ and $\omega_{\max_i}$ for $i \in [0,k]$ yielding a total of $2k$ scalar values.

In the context an input video volume, treated as a collection of $H \times W$ signals of length $N_t$, the procedure extracts the discriminative frequency and its corresponding power at each spatial position. In other words, the spectral decomposition layer performs the mapping $\mathbb{R}^{C \times N_t \times H\times W } \rightarrow \mathbb{R}^{2kC \times H \times W}$. The videos' temporal dimension is squeezed and the result can be considered a multi-channel feature map -- to be further processed by any 2D convolutional layer. We reduce spectral leakage by adopting a Hanning window before the DFT. The batched version of the proposed layer is formalized in \Cref{alg:spectral-decomposition-layer}.

\vspace{2mm}

\noindent \textbf{Embedding Function}. The specification of $s_{\phi}(\*x)$, with the spectral decomposition layer at its core, is illustrated in \Cref{fig:network-architecture}. First, our model convolves the input video $\*x$ with a temporal Gaussian filter followed by two spatially oriented first-order derivative filters. Both resulting video volumes are two-times spatially subsampled by means of max-pooling. Successively, the filtered video representations are fed through the spectral decomposition layer to produce spectral power and frequency maps. The outputs are stacked into a multi-channel feature map to be further processed by a number of 2D convolutional filters with trainable weights $\phi$. We use $3$ standard ResNet blocks \cite{he2016deep} and a final linear layer that maps to the $\mathbb{R}^{D_e}$ embedding space. We refer to our network as \emph{Spectral Decomposition Network (SDN)}.

\vspace{2mm}

\noindent \textbf{Network Details.} Our network is implemented in PyTorch \cite{paszke2017automatic}. We assert that all network inputs are temporally sampled at $25$ fps. After that, we use a temporal Gaussian with $\sigma_t = 1$ and first-order Gaussian derivative filters with $\sigma_{x,y} = 2$. For training the embedding function with the contrastive loss, we adopt a margin of $1$ and use the \emph{Batch All} sampling strategy \cite{hermans2017defense,ding2015deep}. The spectral decomposition layer selects the single most discriminative frequency (\ie $k=1$). Adding secondary frequency peaks to the feature maps did not yield substantial performance gains. The size of our embeddings is fixed ($D_e = 512$) for the paper. Input video clips of size $224 \times 224$ are converted to grayscale. We optimize the weights using ADAM \cite{kingma2015adam} with mini-batches of $32$, learning rate $10^{-2}$ and a weight decay of $2\cdot 10^{-3}$.

%% file: 04c_flags_datasets.tex
\section{Real and Simulated Datasets}
\label{sec:datasets}

\begin{figure}
    \centering
    \begin{minipage}{0.2\columnwidth}
    \centering
        \includegraphics[width=\textwidth,trim={1.15cm 0 1.15cm 0},clip]{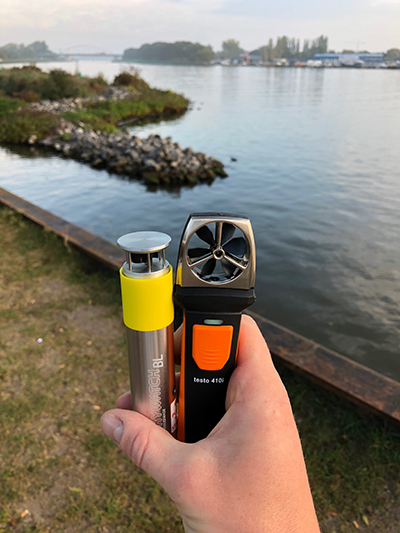}
    \end{minipage}
    \begin{minipage}{0.79\columnwidth}
        \begin{subfigure}{.19\columnwidth}
            \centering
            \includegraphics[width=\textwidth]{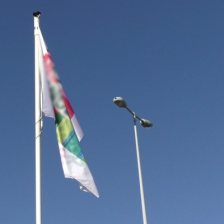}
        \end{subfigure}
        \begin{subfigure}{.19\columnwidth}
            \centering
            \includegraphics[width=\textwidth]{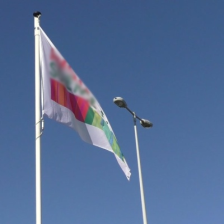}
        \end{subfigure}
        \begin{subfigure}{.19\columnwidth}
            \centering
            \includegraphics[width=\textwidth]{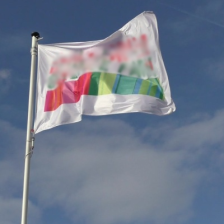}
        \end{subfigure}
        \begin{subfigure}{.19\columnwidth}
            \centering
            \includegraphics[width=\textwidth]{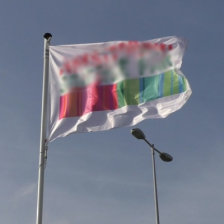}
        \end{subfigure}
        \begin{subfigure}{.19\columnwidth}
            \centering
            \includegraphics[width=\textwidth]{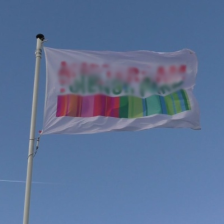}
        \end{subfigure}
        \\[1mm]
        \begin{subfigure}{.19\columnwidth}
            \centering
            \includegraphics[width=\textwidth]{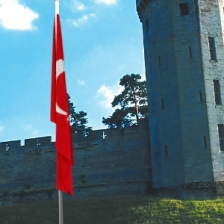}
        \end{subfigure}
        \begin{subfigure}{.19\columnwidth}
            \centering
            \includegraphics[width=\textwidth]{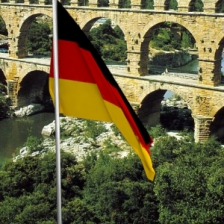}
        \end{subfigure}
        \begin{subfigure}{.19\columnwidth}
            \centering
            \includegraphics[width=\textwidth]{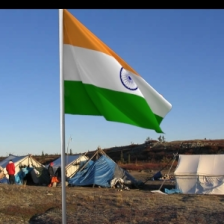}
        \end{subfigure}
        \begin{subfigure}{.19\columnwidth}
            \centering
            \includegraphics[width=\textwidth]{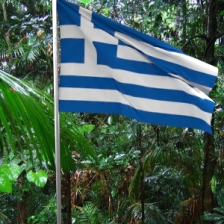}
        \end{subfigure}
        \begin{subfigure}{.19\columnwidth}
            \centering
            \includegraphics[width=\textwidth]{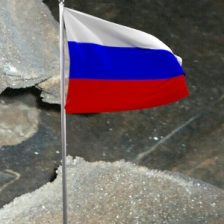}
        \end{subfigure}
    \end{minipage}
    \vspace{-2mm}
    \caption{\emph{Left:} Two anemometers used for gauging the wind speed. \emph{Right top:} Real flag recordings with corresponding wind speeds measured by the anemometer hoisted in the flagpole. \emph{Right bottom:} simulated examples from our FlagSim dataset.} \label{fig:flag-datset-examples}
    \vspace{-5mm}
\end{figure}

\noindent \textbf{Real-world Flag Videos.} To evaluate our method's ability to infer physical parameters from real-world observations, we have set out to collect video recordings of real-world flags with ground-truth wind speed. More precisely, we used two anemometers (\Cref{fig:flag-datset-examples}) to measure the wind speed that exposes the flag. After calibration and verification of the meters, we hoisted one of them in a flag pole at the same height as the flag to ensure accurate and local measurements. A Panasonic HC-V770 camera was used for recording the video. In total, we have acquired more than an hour of video over the course of $5$ days in varying wind and weather conditions. Examples of the videos are displayed in \Cref{fig:flag-datset-examples}.

\vspace{2mm}

\noindent \textbf{FlagSim Dataset.} To train the embedding function $s_{\phi}(\*x)$ as discussed in \Cref{subsec:similarity-function}, we introduce the FlagSim dataset consisting of flag simulations and their rendered animations. We simulate flags by random sampling a set of physical parameters $\btheta$ from \Cref{tab:parameter-search-space} and feed them to ArcSim. We fix the flag's aspect ratio to $2:3$ which is widespread around the world. For each flag simulation, represented as sequence of 3D meshes, we use Blender \cite{blender2018} to render multiple flag animations $\*x_{\text{sim}}^i$ at different render settings $\*\zeta^i$. We position the camera at a varying distance from the flagpole and keep a maximum angle of $15^{\circ}$ between the wind direction and camera axis. From a collection of $12$ countries, we randomly sample a flag texture. Background images are selected from the SUN397 dataset \cite{xiao2010sun}. Each simulation produces $60$ cloth meshes at step size $\Delta T = 0.04$~\si{\second} (\ie 25 fps) which we then render at $300\times 300$ resolution. Following this procedure, we generate $1,000$ mesh sequences and render a total of $14,000$ training examples. We additionally generate validation and test sets of $150/3,800$ and $85/3,500$ mesh sequences/renders respectively. Some examples are visualized in \Cref{fig:flag-datset-examples}.

%% file: 05_experiments.tex
\begin{table}[b]
    \centering
    \ra{1.1}
    \caption{Evaluation of our physical similarity function $s_\phi(\*x)$ for FlagSim test examples. We report triplet accuracies \cite{veit2017conditional}. \label{tab:triplet-accuracies}}
    \vspace{-2mm}
    \small
    \begin{tabular}{llllll}
        \toprule
        Input Frames & $10$ & $20$ & $30$ & $40$ & $50$ \\
        \midrule
        Accuracy & $89.3$ & $92.1$ & $\mathbf{96.3}$ & $90.1$ & $92.4$ \\
        \bottomrule
    \end{tabular}
\end{table}

\begin{figure}
	\centering
    \includegraphics[width=1.0\columnwidth]{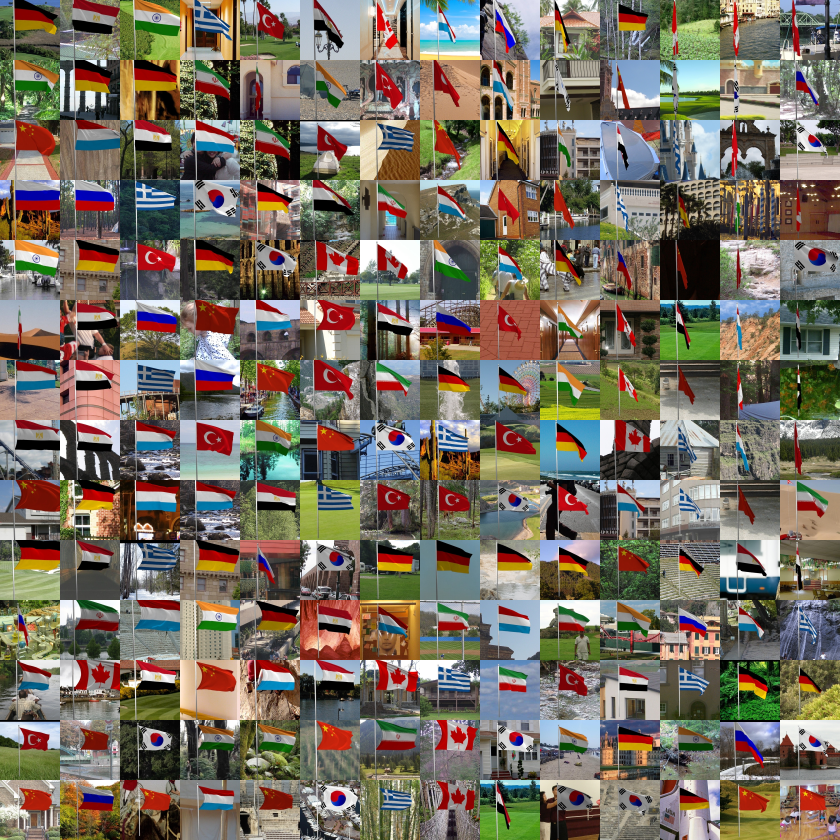}
    \vspace{-5mm}
    \caption{Barnes-Hut t-SNE \cite{van2014accelerating} visualization of the learned flag embedding space. Top-right examples exhibit flags at low wind speeds while bottom-left corresponds to strong winds.} \label{fig:tsne-embeddings}
    \vspace{-4mm}
\end{figure}

\section{Results and Discussion}
\label{sec:experiments}

\noindent \textbf{Physical Similarity Quality.} We first determine the quality of the physical similarity embeddings by measuring how well the network is able to separate examples with similar intrinsic and extrinsic parameters from dissimilar ones. To quantify this, we report the triplet accuracy \cite{veit2017conditional}. We construct $3.5$\si{K} FlagSim triplets from the test set as described in \Cref{subsec:similarity-function}. We consider the SDN trained for video clips of varying number of input frames and report its accuracies in \Cref{tab:triplet-accuracies}. The results demonstrate the effectiveness of our embedding for grouping examples with similar intrinsic and extrinsic parameters $(\btheta_i, \btheta_e)$ on the manifold while mapping dissimilar examples to distant points. We conclude that $30$ input frames is best with a triplet accuracy of $96.3\%$ and therefore use $30$ input frames in the remainder of this paper. We also visualize a subset of the embedding space in \Cref{fig:tsne-embeddings} and observe that simulated flag instances with low wind speeds are clustered in the top-right corner whereas lashing wind speeds live in the bottom-left. 

\vspace{2mm}

\begin{figure*}[t]
    \fboxsep=0mm  %
    \fboxrule=2pt %
    \centering
    \begin{minipage}[c]{0.85\textwidth}
        \centering
        \begin{subfigure}{.19\textwidth}
            \centering
            \fcolorbox{greencustom}{white}{\includegraphics[width=\textwidth,trim={1.2cm 1cm 0.3cm 0.4cm},clip]{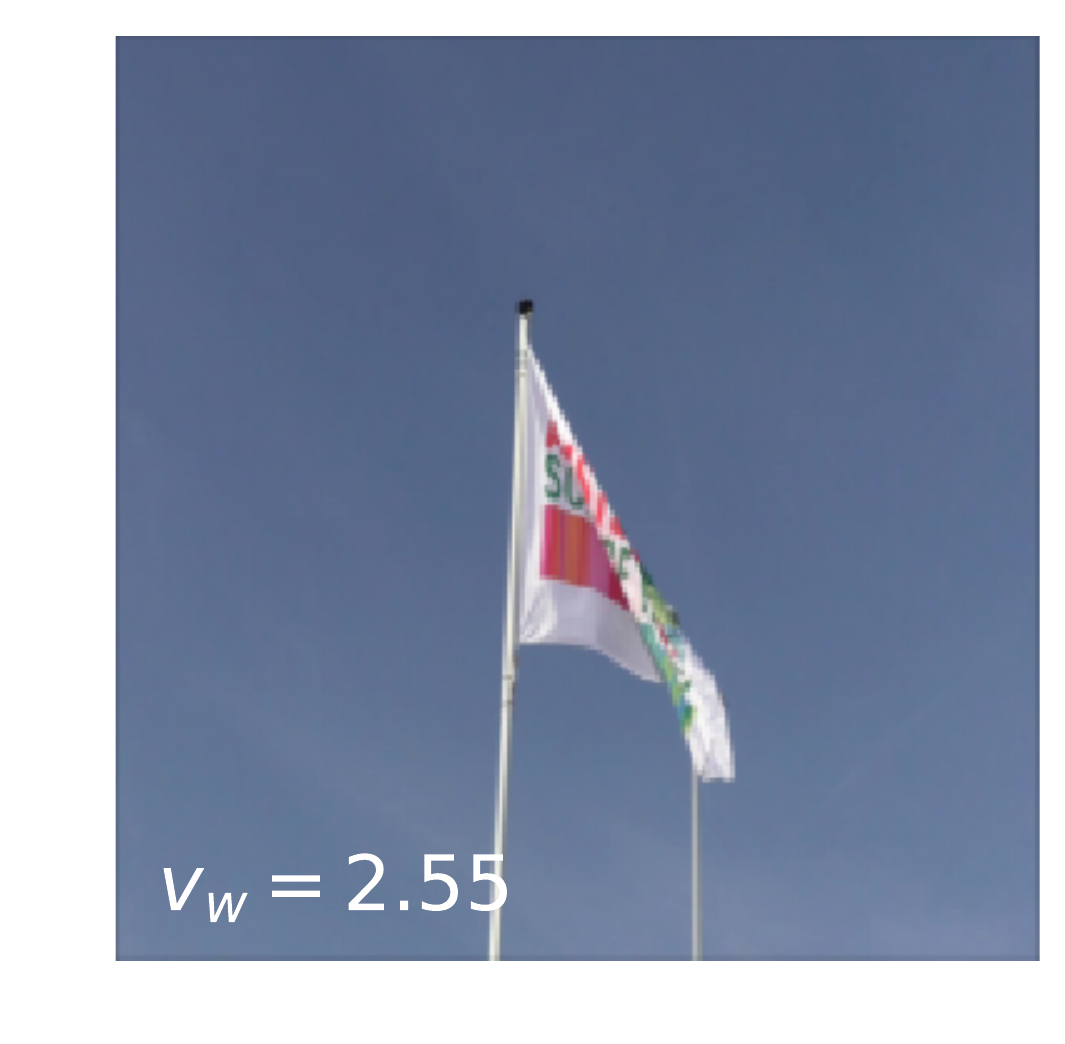}}
        \end{subfigure}
        \hfill
        \begin{subfigure}{.19\textwidth}
            \centering
            \fcolorbox{lightgray}{white}{\includegraphics[width=\textwidth,trim={1.3cm 1.1cm 0.35cm 0.8cm},clip]{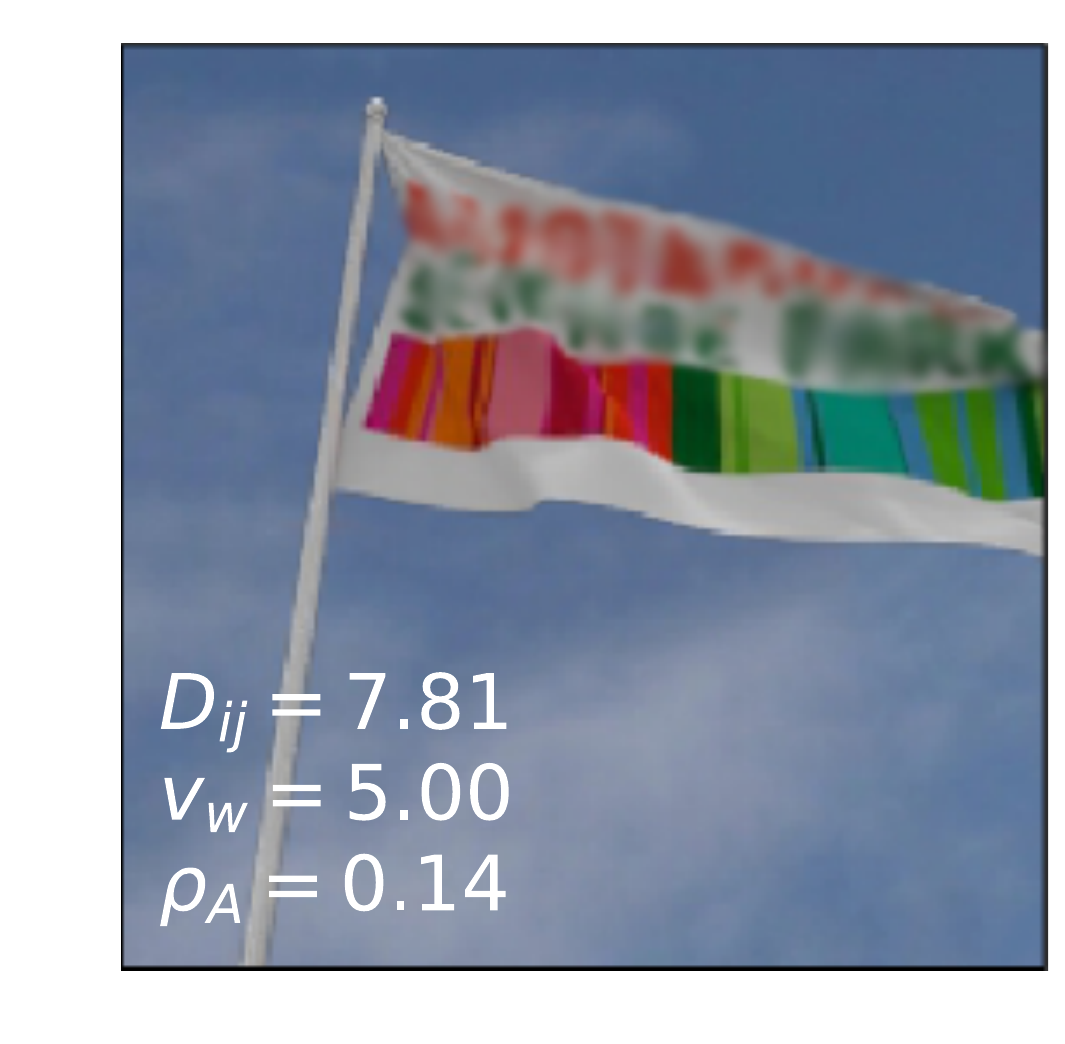}}
        \end{subfigure}
        \hfill
        \begin{subfigure}{.19\textwidth}
            \centering
            \fcolorbox{lightgray}{white}{\includegraphics[width=\textwidth,trim={1.3cm 1.1cm 0.35cm 0.8cm},clip]{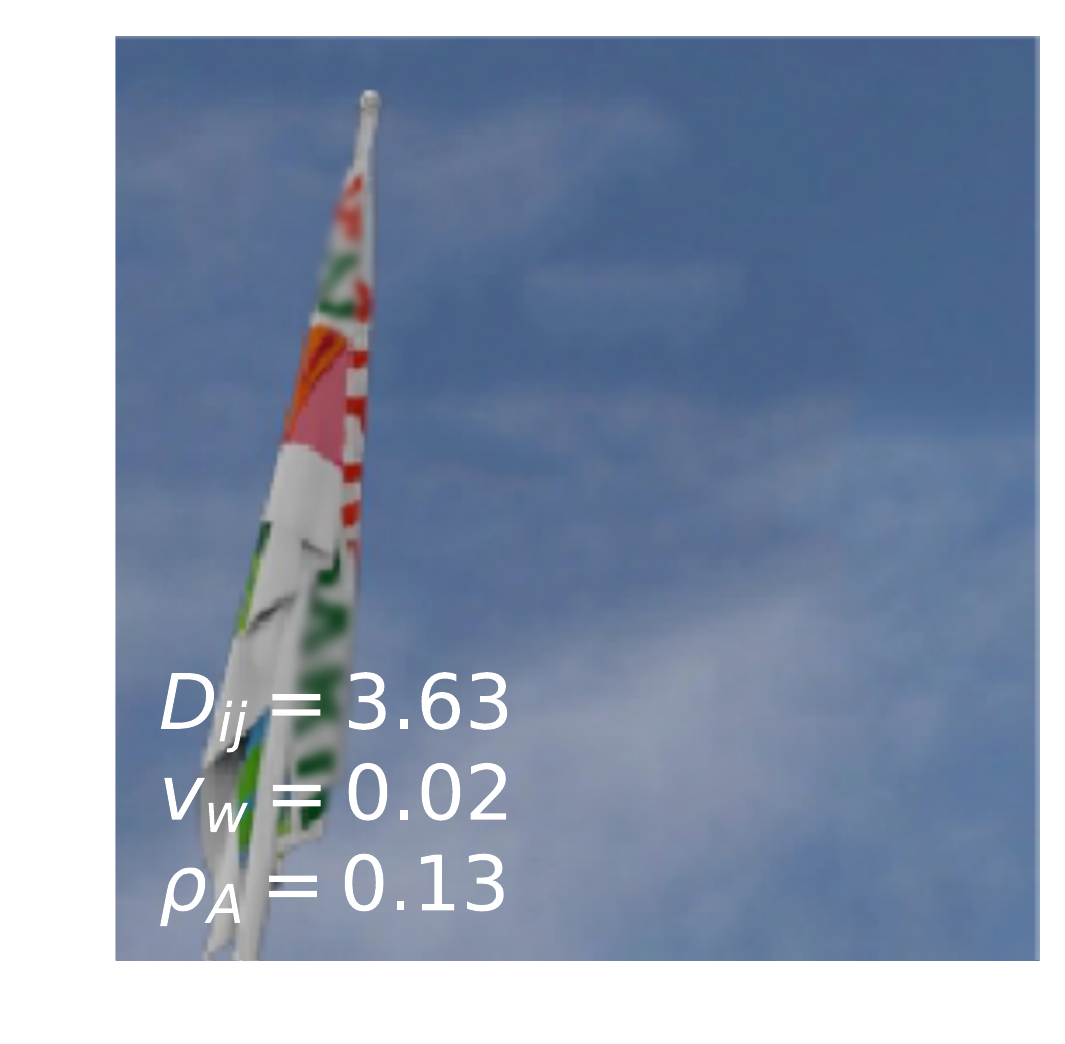}}
        \end{subfigure}
        \hfill
        \begin{subfigure}{.19\textwidth}
            \centering
            \fcolorbox{lightgray}{white}{\includegraphics[width=\textwidth,trim={1.3cm 1.1cm 0.35cm 0.8cm},clip]{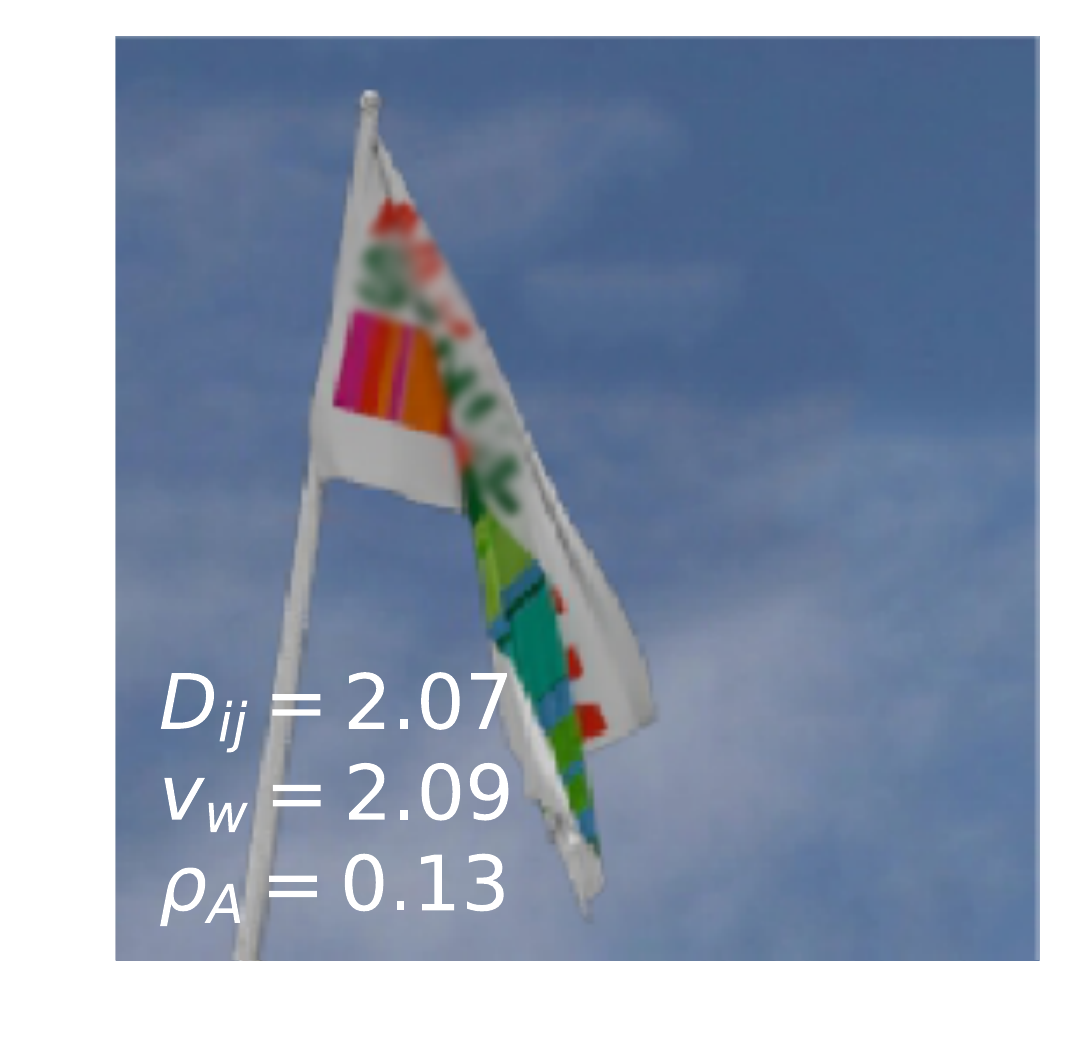}}
        \end{subfigure}
        \hfill
        \begin{subfigure}{.19\textwidth}
            \centering
            \fcolorbox{lightgray}{white}{\includegraphics[width=\textwidth,trim={1.3cm 1.1cm 0.35cm 0.8cm},clip]{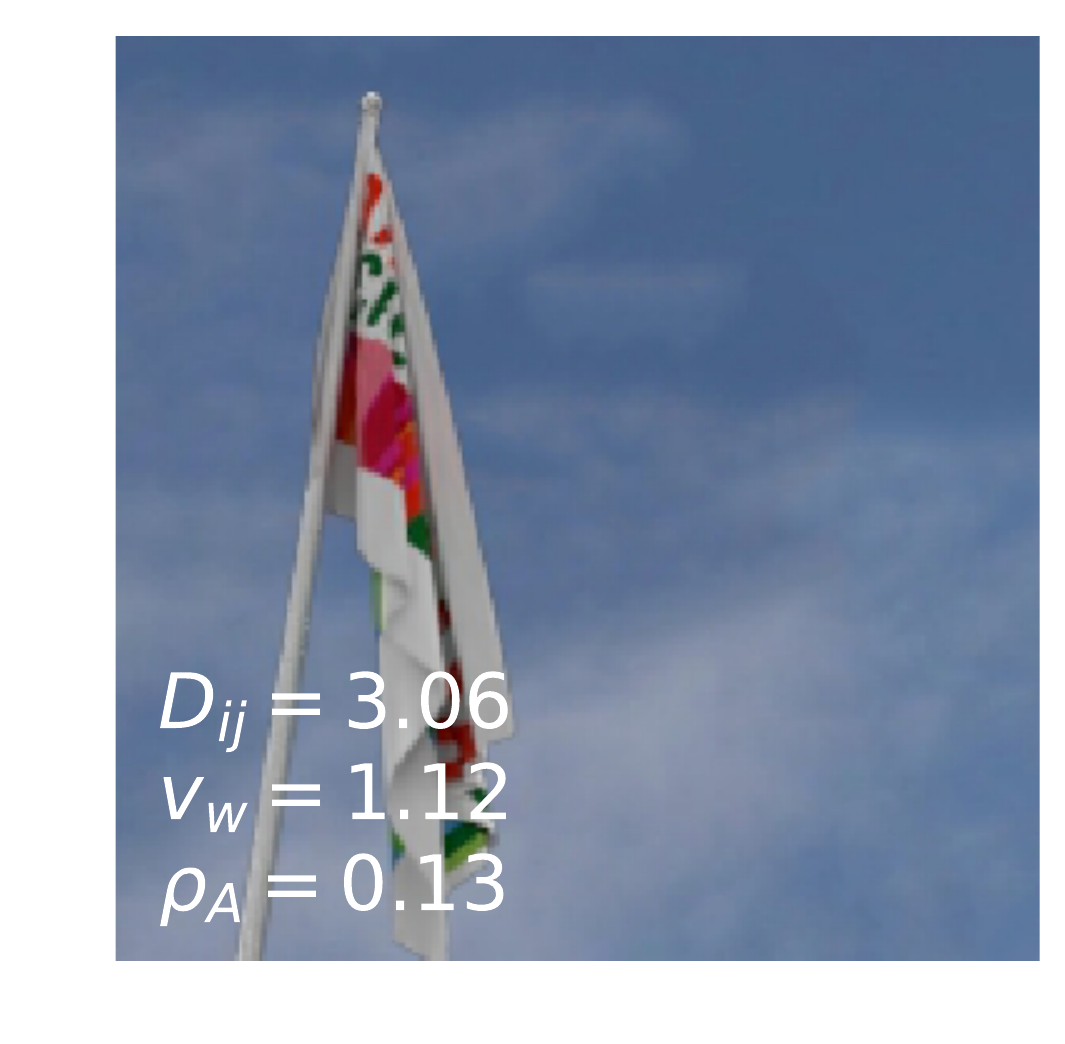}}
        \end{subfigure}
    \end{minipage}
    \\
    \begin{subfigure}{\textwidth}
        \includegraphics[width=\textwidth]{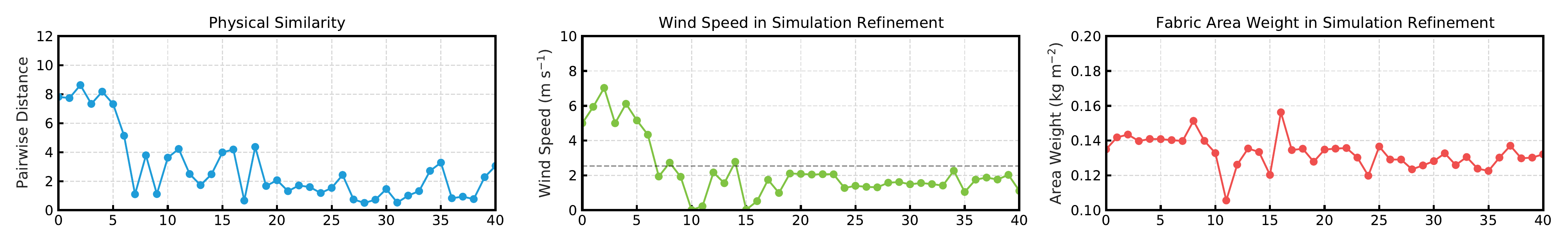}
    \end{subfigure}
    \vspace{-3mm}
    \caption{Result of our iterative simulation refinement for a target video capturing flag in the wind. \emph{Top left:} frame from the real-world target video clip with ground-truth wind speed measured using an anemometer. \emph{Top remaining:} simulated examples throughout the refinement process with corresponding simulation parameters. \emph{Bottom:} development throughout the refinement process for $40$ iteration steps. We plot the distance between simulation and target instance in the embedding space, the wind speed (\ms) and the area weight (\kgmm). In the center plot, we annotate the ground-truth wind speed with a dashed line. The refinement process converges towards the real value.} \label{fig:simrefine_results}
    \vspace{-4mm}
  \end{figure*}

\begin{figure}
	\centering
    \includegraphics[width=\columnwidth,trim={3mm 5mm 3mm 0},clip]{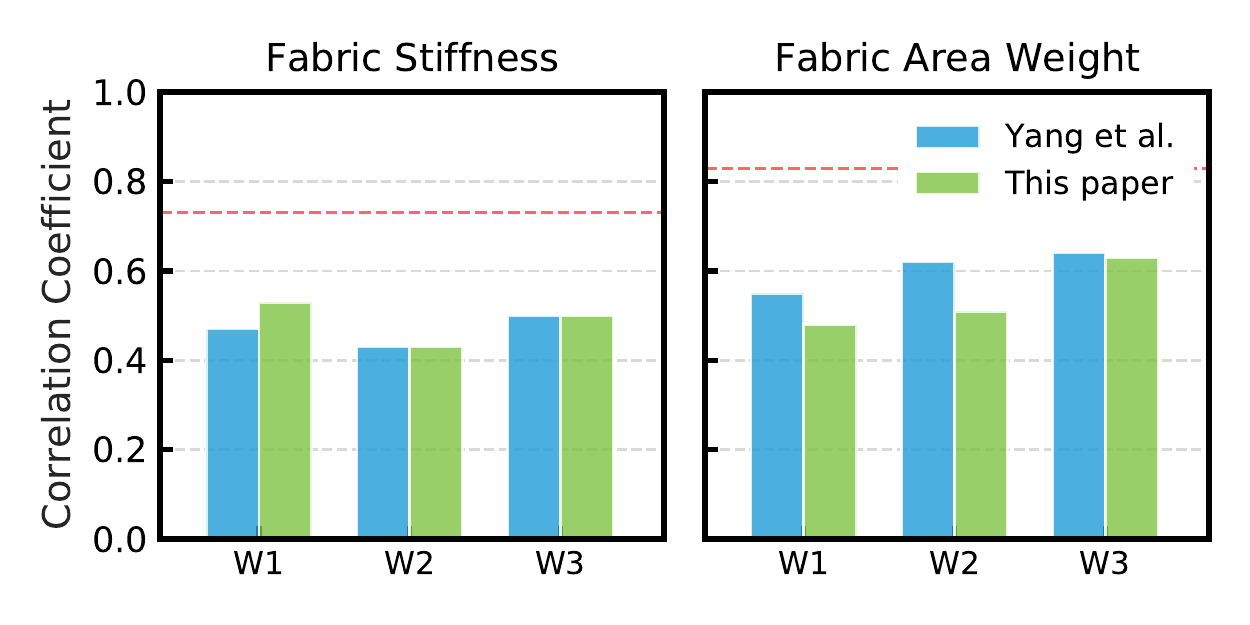}
    \vspace{-5mm}
    \caption{Intrinsic cloth material recovery on real videos. We report the Pearson correlation coefficients (higher is better) between predicted material type and both ground-truth stiffness/density on the Bouman \etal \cite{bouman2013estimating} hanging cloth dataset. The dashed red line indicates human performance as determined by \cite{bouman2013estimating}.} \label{fig:barchart-bouman}
    \vspace{-4mm}
\end{figure}

\noindent \textbf{External Parameter from Simulation.} Next, we compare our SDN against two existing methods on the task of regressing the external model parameter $\btheta_e$, \ie the wind speed ($v_w \in [0, 10]$ \ms{}) on our FlagSim dataset. First, we train the CNN+LSTM model proposed by Yang \etal \cite{yang2017learning} on our dataset with a regression head connected to the LSTM's output. This model was introduced for inferring material properties from video and seems appropriate for the task. We tried both training from scratch and using ImageNet-pretrained weights but found similar performance. Second, we train ResNet-18 \cite{he2016deep} with multiple input frames followed by temporal average pooling of the final activations \cite{karpathy2014large}. We were unable to fit the ResNet model with more than $20$ input frames in GPU memory. After training all methods, we report the mean squared error (MSE) and accuracy within $0.5$ \ms{} (Acc@0.5) in \Cref{tab:flagsim-wind-strength}. While our method has significantly fewer parameters ($2.6${\sc M} versus $11.2${\sc M} and $66.4${\sc M}) and training data is abundant, our SDN outperforms existing methods on estimating the external model parameter.

\vspace{2mm}

\begin{table}
    \centering
    \caption{External wind speed prediction from simulation. We regress the wind speed ($v_w \in \btheta_e$) on our FlagSim dataset. The metrics are computed over the $3.5${\sc{K}} test examples. Target velocities range from $0$~\ms{} (no wind) to $10$~\ms{} (strong wind). \label{tab:flagsim-wind-strength} }
    \ra{1.1}
    \vspace{-2mm}
    \small
    \begin{tabular}{lccc}
        \toprule

        Model & Input Modality & MSE $\downarrow$ & Acc@$0.5$ $\uparrow$ \\ 
        \midrule
        Yang \etal \cite{yang2017learning} & $10\times 227 \times 227$ & $0.380$ & $0.620$ \\ 
        ResNet-18 & $\hphantom{0}1\times 224 \times 224$ & $0.381$ & $0.615$ \\
        ResNet-18 & $10 \times 224 \times 224$  & $0.264$ & $0.734$ \\
        ResNet-18 & $20 \times 224 \times 224$  & $0.207$ & $0.775$ \\
        \midrule
        This paper & $20\times 224 \times 224$ & $0.183$ & $0.813$ \\ 
        This paper & $30\times 224 \times 224$ & $\mathbf{0.180}$ & $\mathbf{0.838}$ \\ 
        \bottomrule
    \end{tabular}
    \vspace{-4mm}
\end{table}

\noindent \textbf{Real-world Intrinsic Cloth Parameter Recovery.} In this experiment, we assess the effectiveness of our SDN for estimating intrinsic cloth material properties from real-world video. We compare against Yang \etal \cite{yang2017learning} on the hanging cloth dataset of Bouman \etal \cite{bouman2013estimating}. Each of the $90$ videos shows one of $30$ cloth types hanging down while being excited by a fan at $3$ wind speeds (W1-3). The goal is to infer the cloth's stiffness and area weight. From our SDN trained on FlagSim with contrastive loss, we extract the embedding vectors for the $90$ videos project them into a $50$-dimensional space using PCA. Then we train a linear regression model using leave-one-out following \cite{bouman2013estimating}. The results are displayed in \Cref{fig:barchart-bouman}. While not outperforming the specialized method of \cite{yang2017learning}, we find that our flag-based features generalize to intrinsic cloth material recovery. This is noteworthy, as our SDN was trained on flags of lightweight materials exhibiting predominantly horizontal motion. %

\vspace{2mm}

\noindent \textbf{Real-world Combined Parameter Refinement.} Our goal is to iteratively refine physics simulation based on real-world observation. Here, we demonstrate the full simulation refinement (\Cref{fig:method-overview}) by optimizing over intrinsic and extrinsic model parameters $(\btheta_i, \btheta_e)$ from real-world flag videos. For a run through the algorithm, we randomly sample a simulation target from our real-world flag recordings. The parameter range of the intrinsic ($16\times$) and extrinsic ($1\times$) is normalized to the domain $[-1,+1]$ and are all initialized $0$, \ie their center values. As our emphasis is not on inferring the render parameters $\*\zeta$ from the real-world observation, we set its values in oracle fashion although the embedding function is moderately robust to this variance (\Cref{fig:tsne-embeddings}). In each step, we simulate the cloth meshes with current parameters $\btheta_i, \btheta_e$ and render its video clip with fixed render parameters $\*\zeta$. Both the simulation and real-world video clip are then mapped onto the embedding space using $s_\phi(\*x)$ and we compute their pairwise distance \eqref{eq:distance-function}. Finally, the Bayesian optimization's acquisition function (\Cref{subsec:parameter-optimization}) determines to where make the next evaluation $\btheta_i, \btheta_e \in [-1,+1]$ to maximize the expected improvement, \ie refining the simulation towards the real-world observation. The next iteration starts by denormalizing the parameters and running the simulation. We run the algorithm for $40$ refinement steps. In \Cref{fig:simrefine_results} we demonstrate our method's development over the course of optimization. Most importantly, we observe a gradual decrease in the pairwise distance between simulation and real-world example, indicating a successful refinement of the simulation parameters towards the actual observation. Furthermore, we note that the wind speed converges towards the ground-truth wind speed as indicated with a dashed line.

%% file: 06_conclusion.tex
\section{Conclusion}
\label{sec:conclusion}

We have presented perception-based refinement of intrinsic and extrinsic physical model parameters. The optimization is guided by the similarity between current simulation and a real-world observation. By leveraging simulated data, we have introduced the training of a physical similarity function. This enables quantifying the physical correspondence between real and simulated data. As specific instance of our method, we have discussed flags in the wind. In this context, we have introduced the SDN that performs temporal spectral decomposition of a video volume.

\ifcvprfinal
    \noindent \textbf{Acknowledgements.} We would like to thank Rik Holsheimer for his help with the real-world flag dataset acquisition. 
\fi

%% file: ms.bbl
\begin{thebibliography}{10}\itemsep=-1pt

\bibitem{baraff1998large}
David Baraff and Andrew Witkin.
\newblock Large steps in cloth simulation.
\newblock In {\em SIGGRAPH}, 1998.

\bibitem{baraff2003untangling}
David Baraff, Andrew Witkin, and Michael Kass.
\newblock Untangling cloth.
\newblock In {\em SIGGRAPH}, 2003.

\bibitem{batchelor1967introduction}
CK Batchelor and GK Batchelor.
\newblock {\em An introduction to fluid dynamics}.
\newblock Cambridge university press, 1967.

\bibitem{bergstra2012random}
James Bergstra and Yoshua Bengio.
\newblock Random search for hyper-parameter optimization.
\newblock {\em JMLR}, 13(Feb):281--305, 2012.

\bibitem{bhat2003estimating}
Kiran~S Bhat, Christopher~D Twigg, Jessica~K Hodgins, Pradeep~K Khosla, Zoran
  Popovi{\'c}, and Steven~M Seitz.
\newblock Estimating cloth simulation parameters from video.
\newblock In {\em SIGGRAPH}, 2003.

\bibitem{blender2018}
{Blender Online Community}.
\newblock {\em Blender - a 3D modelling and rendering package}.
\newblock Blender Foundation, 2018.

\bibitem{bouman2013estimating}
Katherine~L Bouman, Bei Xiao, Peter Battaglia, and William~T Freeman.
\newblock Estimating the material properties of fabric from video.
\newblock In {\em ICCV}, 2013.

\bibitem{bridson2005simulation}
Robert Bridson, Sebastian Marino, and Ronald Fedkiw.
\newblock Simulation of clothing with folds and wrinkles.
\newblock In {\em SIGGRAPH}, 2005.

\bibitem{bromley1994signature}
Jane Bromley, Isabelle Guyon, Yann LeCun, Eduard S{\"a}ckinger, and Roopak
  Shah.
\newblock Signature verification using a" siamese" time delay neural network.
\newblock In {\em NIPS}, 1994.

\bibitem{cardona2019seeing}
Jennifer~L Cardona, Michael~F Howland, and John~O Dabiri.
\newblock Seeing the wind: Visual wind speed prediction with a coupled
  convolutional and recurrent neural network.
\newblock {\em arXiv preprint arXiv:1905.13290}, 2019.

\bibitem{craik1967nature}
Kenneth James~Williams Craik.
\newblock {\em The Nature of Explanation}, volume 445.
\newblock CUP Archive, 1967.

\bibitem{davis2015visual}
Abe Davis, Katherine~L Bouman, Justin~G Chen, Michael Rubinstein, Fredo Durand,
  and William~T Freeman.
\newblock Visual vibrometry: Estimating material properties from small motion
  in video.
\newblock In {\em CVPR}, 2015.

\bibitem{ding2015deep}
Shengyong Ding, Liang Lin, Guangrun Wang, and Hongyang Chao.
\newblock Deep feature learning with relative distance comparison for person
  re-identification.
\newblock {\em Pattern Recognition}, 48(10):2993--3003, 2015.

\bibitem{eloy2008aeroelastic}
Christophe Eloy, Romain Lagrange, Claire Souilliez, and Lionel Schouveiler.
\newblock Aeroelastic instability of cantilevered flexible plates in uniform
  flow.
\newblock {\em Journal of Fluid Mechanics}, 611:97--106, 2008.

\bibitem{haddon1998shading}
John Haddon and David Forsyth.
\newblock Shading primitives: Finding folds and shallow grooves.
\newblock In {\em ICCV}, 1998.

\bibitem{hadsell2006dimensionality}
Raia Hadsell, Sumit Chopra, and Yann LeCun.
\newblock Dimensionality reduction by learning an invariant mapping.
\newblock In {\em CVPR}, 2006.

\bibitem{he2016deep}
Kaiming He, Xiangyu Zhang, Shaoqing Ren, and Jian Sun.
\newblock Deep residual learning for image recognition.
\newblock In {\em CVPR}, 2016.

\bibitem{hegarty2004mechanical}
Mary Hegarty.
\newblock Mechanical reasoning by mental simulation.
\newblock {\em Trends in cognitive sciences}, 8(6):280--285, 2004.

\bibitem{hermans2017defense}
Alexander Hermans, Lucas Beyer, and Bastian Leibe.
\newblock In defense of the triplet loss for person re-identification.
\newblock {\em arXiv preprint arXiv:1703.07737}, 2017.

\bibitem{huang2010three}
Wei-Xi Huang and Hyung~Jin Sung.
\newblock 3d simulation of a flapping flag in a uniform flow.
\newblock {\em Journal of Fluid Mechanics}, 653:301--336, 2010.

\bibitem{karpathy2014large}
Andrej Karpathy, George Toderici, Sanketh Shetty, Thomas Leung, Rahul
  Sukthankar, and Li Fei-Fei.
\newblock Large-scale video classification with convolutional neural networks.
\newblock In {\em CVPR}, 2014.

\bibitem{kingma2015adam}
Diederik~P Kingma and Jimmy Ba.
\newblock Adam: A method for stochastic optimization.
\newblock In {\em ICLR}, 2015.

\bibitem{li2016fall}
Wenbin Li, Seyedmajid Azimi, Ale{\v{s}} Leonardis, and Mario Fritz.
\newblock To fall or not to fall: A visual approach to physical stability
  prediction.
\newblock {\em arXiv preprint arXiv:1604.00066}, 2016.

\bibitem{lindeberg2018dense}
Tony Lindeberg.
\newblock Dense scale selection over space, time, and space-time.
\newblock {\em SIAM Journal on Imaging Sciences}, 11(1):407--441, 2018.

\bibitem{meka2018lime}
Abhimitra Meka, Maxim Maximov, Michael Zollhoefer, Avishek Chatterjee,
  Hans-Peter Seidel, Christian Richardt, and Christian Theobalt.
\newblock Lime: Live intrinsic material estimation.
\newblock In {\em CVPR}, 2018.

\bibitem{mottaghi2016newtonian}
Roozbeh Mottaghi, Hessam Bagherinezhad, Mohammad Rastegari, and Ali Farhadi.
\newblock Newtonian scene understanding: Unfolding the dynamics of objects in
  static images.
\newblock In {\em CVPR}, 2016.

\bibitem{mottaghi2016happens}
Roozbeh Mottaghi, Mohammad Rastegari, Abhinav Gupta, and Ali Farhadi.
\newblock ``what happens if...'' learning to predict the effect of forces in
  images.
\newblock In {\em ECCV}, 2016.

\bibitem{narain2012adaptive}
Rahul Narain, Armin Samii, and James~F O'Brien.
\newblock Adaptive anisotropic remeshing for cloth simulation.
\newblock In {\em SIGGRAPH}, 2012.

\bibitem{oh2018bock}
ChangYong Oh, Efstratios Gavves, and Max Welling.
\newblock Bock: Bayesian optimization with cylindrical kernels.
\newblock In {\em ICML}, 2018.

\bibitem{oppenheim1999discrete}
Alan~V Oppenheim.
\newblock {\em Discrete-time signal processing}.
\newblock Pearson Education India, 1999.

\bibitem{paszke2017automatic}
Adam Paszke, Sam Gross, Soumith Chintala, Gregory Chanan, Edward Yang, Zachary
  DeVito, Zeming Lin, Alban Desmaison, Luca Antiga, and Adam Lerer.
\newblock Automatic differentiation in pytorch.
\newblock In {\em NIPS-W}, 2017.

\bibitem{provot1995deformation}
Xavier Provot et~al.
\newblock Deformation constraints in a mass-spring model to describe rigid
  cloth behaviour.
\newblock In {\em Graphics interface}. Canadian Information Processing Society,
  1995.

\bibitem{runia2018repetition}
Tom Frederik~Hugo Runia, Cees G~M Snoek, and Arnold W~M Smeulders.
\newblock Repetition estimation.
\newblock {\em arXiv preprint arXiv:1806.06984}, 2018.

\bibitem{sakaino2008fluid}
Hidetomo Sakaino.
\newblock Fluid motion estimation method based on physical properties of waves.
\newblock In {\em CVPR}, 2008.

\bibitem{shelley2011flapping}
Michael~J Shelley and Jun Zhang.
\newblock Flapping and bending bodies interacting with fluid flows.
\newblock {\em Annual Review of Fluid Mechanics}, 43:449--465, 2011.

\bibitem{slaughter2012linearized}
William~S Slaughter.
\newblock {\em The linearized theory of elasticity}.
\newblock Springer Science \& Business Media, 2012.

\bibitem{snoek2012practical}
Jasper Snoek, Hugo Larochelle, and Ryan~P Adams.
\newblock Practical bayesian optimization of machine learning algorithms.
\newblock In {\em NIPS}, 2012.

\bibitem{spencer2004water}
Lisa Spencer and Mubarak Shah.
\newblock Water video analysis.
\newblock In {\em ICIP}, 2004.

\bibitem{sun2003video}
Meng Sun, Allan~D Jepson, and Eugene Fiume.
\newblock Video input driven animation (vida).
\newblock In {\em ICCV}, 2003.

\bibitem{taneda1968waving}
Sadatoshi Taneda.
\newblock Waving motions of flags.
\newblock {\em Journal of the Physical Society of Japan}, 24(2):392--401, 1968.

\bibitem{tian2013role}
Fang-Bao Tian.
\newblock Role of mass on the stability of flag/flags in uniform flow.
\newblock {\em Applied Physics Letters}, 103(3):034101, 2013.

\bibitem{van2014accelerating}
Laurens Van Der~Maaten.
\newblock Accelerating t-sne using tree-based algorithms.
\newblock {\em JMLR}, 15(1):3221--3245, 2014.

\bibitem{veit2017conditional}
Andreas Veit, Serge Belongie, and Theofanis Karaletsos.
\newblock Conditional similarity networks.
\newblock In {\em CVPR}, 2017.

\bibitem{wang2011data}
Huamin Wang, James~F O'Brien, and Ravi Ramamoorthi.
\newblock Data-driven elastic models for cloth: modeling and measurement.
\newblock In {\em SIGGRAPH}, 2011.

\bibitem{wejchert1991animation}
Jakub Wejchert and David Haumann.
\newblock Animation aerodynamics.
\newblock In {\em SIGGRAPH}, 1991.

\bibitem{white2007capturing}
Ryan White, Keenan Crane, and David~A Forsyth.
\newblock Capturing and animating occluded cloth.
\newblock In {\em SIGGRAPH}, 2007.

\bibitem{wu2016physics}
Jiajun Wu, Joseph~J Lim, Hongyi Zhang, Joshua~B Tenenbaum, and William~T
  Freeman.
\newblock Physics 101: Learning physical object properties from unlabeled
  videos.
\newblock In {\em BMVC}, 2016.

\bibitem{wu2015galileo}
Jiajun Wu, Ilker Yildirim, Joseph~J Lim, Bill Freeman, and Josh Tenenbaum.
\newblock Galileo: Perceiving physical object properties by integrating a
  physics engine with deep learning.
\newblock In {\em NIPS}, 2015.

\bibitem{xiao2010sun}
Jianxiong Xiao, James Hays, Krista~A Ehinger, Aude Oliva, and Antonio Torralba.
\newblock Sun database: Large-scale scene recognition from abbey to zoo.
\newblock In {\em CVPR}, 2010.

\bibitem{xue2018seeing}
Tianfan Xue, Jiajun Wu, Zhoutong Zhang, Chengkai Zhang, Joshua~B Tenenbaum, and
  William~T Freeman.
\newblock Seeing tree structure from vibration.
\newblock In {\em ECCV}, 2018.

\bibitem{yang2017learning}
Shan Yang, Junbang Liang, and Ming~C Lin.
\newblock Learning-based cloth material recovery from video.
\newblock In {\em ICCV}, 2017.

\bibitem{yang2018physics}
Shan Yang, Zherong Pan, Tanya Amert, Ke Wang, Licheng Yu, Tamara Berg, and
  Ming~C Lin.
\newblock Physics-inspired garment recovery from a single-view image.
\newblock {\em ACM Transactions on Graphics (TOG)}, 37(5):170, 2018.

\end{thebibliography}
